\documentclass{article}
\usepackage{times,makeidx,psfig,a4wide}

\def\Aixi{{\sc Aixi} }
\def\Aixin{{\sc Aixi}}

\def\gm{G\"{o}del machine }

\def\gmn{G\"{o}del machine}

\begin{document}
\title{New Millennium AI and the Convergence of History}
\date{}
\author{J\"{u}rgen Schmidhuber \\
TU Munich, Boltzmannstr. 3,  85748 Garching, M\"{u}nchen, Germany \& \\
IDSIA, Galleria 2, 6928 Manno (Lugano), Switzerland \\
{\tt juergen@idsia.ch - http://www.idsia.ch/\~{ }juergen}}

\maketitle

\begin{abstract}
Artificial Intelligence (AI) has recently become a real formal science:
the new millennium brought the first mathematically sound, asymptotically
optimal, universal problem solvers, providing a new, rigorous foundation for the
previously largely heuristic field of General AI and embedded agents.
At the same time there has been rapid progress
in practical methods for learning true sequence-processing
programs, as opposed to traditional methods limited to
stationary pattern association. Here we will briefly review some of
the new results, and speculate about future developments, pointing
out that the time intervals between the most notable events
in over 40,000 years or $2^9$ lifetimes of 
human history have sped up exponentially, 
apparently converging to zero within the next few decades.
Or is this impression just a by-product of the way humans allocate
memory space to past events?
\end{abstract}

%\tableofcontents

\section{Introduction}
\label{introduction}

In 2003 we observed \cite{Schmidhuber:03newai1,Schmidhuber:03speedup}
that each major breakthrough
in computer science tends to come roughly twice as fast 
as the previous one, roughly matching a century-based scale: 
In 1623 the computing age started with the
first mechanical calculator by Wilhelm Schickard 
(followed by machines of Pascal, 1640, and Leibniz, 1670).
Roughly two centuries later Charles Babbage  came up with the
concept of a program-controlled computer (1834-1840).  
One century later, in 1931, Kurt G\"{o}del layed the foundations of 
theoretical computer science with his
work on universal formal languages and the limits
of proof and computation. His results and Church's extensions thereof
were reformulated by Turing in 1936, while Konrad Zuse built
the first working program-controlled computers (1935-1941), 
using the binary system of Leibniz (1701) 
instead of the more cumbersome decimal system used by Babbage and many others.  
By then all the main ingredients of `modern' computer science were in place.
The next 50 years saw many less radical theoretical advances as well
as faster and faster switches---relays were replaced by tubes by
single transistors by numerous transistors etched on chips---but
arguably this was rather predictable, incremental progress without
earth-shaking events.
Half a century later, however, Berners-Lee triggered the most recent
world-changing development by creating the 
World Wide Web (1990).  

Extrapolating the trend, we should expect the next radical change to
manifest itself one quarter of a century after the most recent one,
that is, by 2015, when some computers will 
already match brains in terms of raw computing power,
according to frequent estimates based on Moore's law, which
suggests a speed-up factor of roughly 1000 per decade, give or take a few years.  
The remaining series of faster and faster additional revolutions should
converge in an {\em Omega point} (term coined by Pierre Teilhard de Chardin, 1916) 
expected between 2030 and 2040, when
individual machines will already approach the raw computing power of all human
brains combined (provided Moore's law does not break down---compare Stanislaw 
Ulam's concept of an approaching {\em historic singularity} (cited
in \cite{Kurzweil:05}) or Vinge's closely related technological singularity \cite{Vinge:93}
as well as the subsequent speculations of Moravec \cite{Moravec:99} and Kurzweil \cite{Kurzweil:05}).
Many of the present readers of this article should still be alive then.

Will the software and the theoretical advances
keep up with the hardware development?
We are convinced they will. In fact, the new millennium has
already brought fundamental new insights into the problem
of constructing theoretically optimal rational agents or 
universal Artificial Intelligences (AIs, more on this below).
On the other hand, on a more practical level, there has been rapid progress
in learning algorithms for agents interacting with a dynamic environment, 
autonomously discovering true sequence-processing,
problem-solving programs, as opposed to the
reactive mappings from stationary inputs to outputs studied in
most traditional machine learning (ML) research.
In what follows, we will briefly review some of the new results, 
then come back to the issue of whether or not history is about to ``converge.''

\section{Overview}
\label{overview}

Since virtually all realistic sensory inputs of robots and other
cognitive systems are sequential by nature, the future of machine 
learning and AI in general
lies in sequence processing as opposed to processing
of stationary input patterns.
Most traditional methods for learning time series and mappings from
sequences to sequences, however, are based on simple time windows:
one of the numerous feedforward ML techniques such as
feedforward neural nets (NN) \cite{Bishop:95}
or support vector machines \cite{Vapnik:95} is used to map
a restricted, fixed time window of sequential input values to
desired target values.  Of
course such approaches are bound to fail if
there are temporal dependencies exceeding the time window size.
Large time windows, on the other hand, yield
unacceptable numbers of free parameters.

If we want to narrow the gap between learning abilities of humans and
machines, then we will have to study how
to learn general algorithms instead of such reactive mappings.
In what follows we will first discuss very recent {\em universal} 
program learning methods that are optimal in various mathematical senses.
For several reasons, however, these methods are not (yet) practically feasible.
Therefore we will also discuss recent less universal
but more feasible program learners
based on recurrent neural networks.

Finally we will return to the introduction's
topic of exponential speed-up,
extending it to all of human history since the
appearance of the Cro Magnon man roughly 40,000 years ago.

\section{Notation}
\label{notation}

Consider a learning robotic agent with a single life which
consists of discrete cycles or time steps $t=1, 2, \ldots, T$.
Its total lifetime $T$ may or may not be known in advance.
In what follows,the value of any time-varying variable $Q$
at time $t$ ($1 \leq t \leq T$) will be denoted by $Q(t)$,
the ordered sequence of values $Q(1),\ldots,Q(t)$ by $Q(\leq t)$,
and the (possibly empty) sequence $Q(1),\ldots,Q(t-1)$ by $Q(< t)$.

At any given $t$  the robot receives a real-valued input vector $x(t)$ from
the environment and executes a real-valued 
action $y(t)$ which may affect future inputs; at times $t<T$ its goal
is to maximize future success or {\em utility}
\begin{equation}
\label{u}
u(t) =
E_{\mu} \left [ \sum_{\tau=t+1}^T  
r(\tau)~~ \Bigg| ~~ h(\leq t) \right ],
\end{equation}
where $r(t)$ is an additional real-valued reward input at time $t$,
$h(t)$ the ordered triple $[x(t), y(t), r(t)]$
(hence $h(\leq t)$ is the known history up to $t$),
and $E_{\mu}(\cdot \mid \cdot)$ denotes the conditional expectation operator
with respect to some possibly unknown distribution $\mu$ from a set $M$
of possible distributions. Here $M$ reflects
whatever is known about the possibly probabilistic reactions
of the environment.  For example, $M$ may contain all computable
distributions \cite{Solomonoff:64,Solomonoff:78,LiVitanyi:97,Hutter:04book+}.
Note that unlike in most previous work by others \cite{Kaelbling:96,Sutton:98},
but like in much of the author's own previous
work  
\cite{Schmidhuber:97ssa,Schmidhuber:03gm},
there is just one life, no need for predefined repeatable trials, 
no restriction to Markovian 
interfaces between sensors and environment \cite{Schmidhuber:91nips},
and the utility function implicitly takes into account the 
expected remaining lifespan $E_{\mu}(T \mid  h(\leq t))$
and thus the possibility to extend it through appropriate actions
\cite{Schmidhuber:03gm,Schmidhuber:05icann,Schmidhuber:05gmai,Schmidhuber:05gmconscious}.

\section{Universal But Incomputable AI}
\label{unilearn}

Solomonoff's theoretically optimal
universal predictors and their Bayesian learning algorithms  
\cite{Solomonoff:64,Solomonoff:78,LiVitanyi:97,Hutter:04book+}
only assume that the reactions of the environment are sampled from 
an unknown probability distribution $\mu$ contained in a set $M$
of all enumerable distributions---compare text after equation (\ref{u}).
That is, given an observation sequence $q(\leq t)$,
we only assume there exists a computer program
that can compute the probability of the next possible $q(t+1)$, given 
$q(\leq t)$. Since we typically do not know this program,
we predict using a mixture distribution 
\begin{equation}
\label{xi}
\xi(q(t+1)\mid q(\leq t)) =\sum_i w_i\mu_i (q(t+1)\mid q(\leq t)),
\end{equation}
a weighted sum of {\em all} distributions $\mu_i \in \cal M$, $i=1, 2, \ldots$, 
where the sum of the constant weights satisfies $\sum_i w_i \leq 1$. 
It turns out that this is indeed
the best one can possibly do, in a very general sense
\cite{Solomonoff:78,Hutter:04book+}. The drawback
is that the scheme is incomputable, since $M$ contains
infinitely many distributions.

One can increase the theoretical power of the scheme by
augmenting $M$ by certain non-enumerable but
limit-computable distributions \cite{Schmidhuber:02ijfcs},
or restrict it such that it becomes computable,
e.g., by assuming the world is computed
by some unknown but deterministic computer
program sampled from the Speed Prior \cite{Schmidhuber:02colt} which assigns
low probability to environments that are hard to compute by any method.
Under the Speed Prior 
the cumulative a priori probability of all data 
whose computation through an optimal algorithm requires 
more than $O(n)$ resources is $1/n$.  

Can we use the optimal predictors to build an optimal AI?
Indeed, in the new millennium it was shown we can.
At any time $t$, the recent theoretically optimal
yet uncomputable RL algorithm \Aixi \cite{Hutter:04book+} 
uses Solomonoff's universal prediction scheme  
to select those action sequences that promise maximal
future reward up to some horizon, typically  $2t$,
given the current data $h(\leq t)$.
One may adapt this to the case of any finite horizon $T$. That is, in cycle $t+1$, \Aixi
selects as its next action the first action of an action sequence
maximizing $\xi$-predicted reward up to the horizon, appropriately
generalizing eq. (\ref{xi}).
Recent work \cite{Hutter:04book+} demonstrated \Aixin's optimal
use of observations as follows.  The Bayes-optimal policy $p^\xi$ based on
the mixture $\xi$ 
is self-optimizing in the sense that its average
utility value converges asymptotically for all $\mu \in \cal M$ to the
optimal value achieved by the (infeasible) Bayes-optimal policy $p^\mu$
which knows $\mu$ in advance.  The necessary condition that $\cal M$
admits self-optimizing policies is also sufficient.
Furthermore, $p^\xi$ is Pareto-optimal
in the sense that there is no other policy yielding higher or equal
value in {\em all} environments $\nu \in \cal M$ and a strictly higher
value in at least one \cite{Hutter:04book+}.

What are the implications?
The first 50 years of attempts at ``general AI'' 
have been dominated by heuristic approaches 
\cite{Newell:63,SOAR:93,Utgoff:86,Mitchell:97}.
Traditionally many theoretical computer scientists have regarded
the field with contempt for its lack of hard theoretical results.
Things have changed, however.
Although the universal approach above is practically infeasible
due to the incomputability of Solomonoff's prior,
it does provide, for the first time, a mathematically sound theory 
of AI and optimal decision making based on experience, identifying the limits of
both human and artificial intelligence, and providing a yardstick
for any future approach to general AI.

Using such results one can also come up with theoretically optimal ways of 
improving the predictive world model of
a curious robotic agent \cite{Schmidhuber:06cs}. 
The rewards of an optimal reinforcement learner are the predictor's
improvements on the observation history so far.
They encourage the 
reinforcement learner to produce action sequences
that cause the creation and the learning of new, 
previously unknown regularities in the sensory input stream.
It turns out that art and creativity can be explained as by-products of such
intrinsic curiosity rewards:  good observer-dependent 
art deepens the observer's insights about this world or
possible worlds, connecting previously disconnected patterns 
in an initially surprising way that eventually becomes known and boring. 
While previous attempts at describing what is satisfactory 
art or music were informal,  this work
permits the first {\em technical, formal}
approach to understanding the nature of 
art and creativity \cite{Schmidhuber:06cs}. 

Using the Speed Prior mentioned above,
one can scale the universal approach down such 
that it becomes computable  \cite{Schmidhuber:02colt}.
In what follows we will mention ways of introducing additional 
optimality criteria that take into account the computational
costs of prediction and decision making.

\section{Asymptotically Optimal General Problem Solver}
\label{fast}

To take computation time into account in a general,
optimal way \cite{Levin:73} \cite[p. 502-505]{LiVitanyi:97},
the recent asymptotically
optimal search algorithm for {\em all} well-defined problems
\cite{Hutter:01fast+}  allocates part of the total
search time to searching the space of proofs for provably correct
candidate programs with provable upper runtime bounds;
at any given time it
focuses resources on those programs with the currently
best proven time bounds.
The method is as fast as the initially
unknown fastest problem solver for the given problem class,
save for a constant slowdown factor of at most
$1 + \epsilon$, $\epsilon >0$, and an additive
problem class-specific constant.
Unfortunately, however,  the latter may be huge.

Practical applications may not ignore the constants though.
This motivates the next section which
addresses all kinds of optimality
(not just asymptotic optimality).

\section{Optimal Self-Referential General Problem Solver}
\label{gm}

The recent \gmn s 
\cite{Schmidhuber:03gm,Schmidhuber:05icann,Schmidhuber:05gmai,Schmidhuber:05gmconscious}
represent the first class of mathematically rigorous, general, fully
self-referential, self-improving, optimally efficient problem solvers.
In particular, they are applicable to the problem embodied by 
objective (\ref{u}), which obviously does not
care for asymptotic optimality.

The initial software  $\cal S$ of such a \gm contains
an initial problem solver, e.g., one of
the approaches above \cite{Hutter:04book+} or some less general,
typical sub-optimal  method
\cite{Kaelbling:96,Sutton:98}.
Simultaneously, it contains an initial proof
searcher (possibly based on an online variant of Levin's
{\em Universal Search} \cite{Levin:73})
which is used to run and test {\em proof techniques}. The latter
are programs written in a universal programming language implemented
on the \gm within $\cal S$, able to compute proofs
concerning the system's own future performance, based on an axiomatic 
system $\cal A$ encoded in $\cal S$.
$\cal A$ describes the formal {\em utility} function, in our case eq. (\ref{u}),
the hardware properties, axioms of arithmetics and probability
theory and string manipulation etc, and  $\cal S$ itself, which is possible
without introducing circularity
\cite{Schmidhuber:03gm}.

Inspired by Kurt G\"{o}del's celebrated self-referential formulas (1931),
the \gm rewrites any part of its own code in a computable way through
a self-generated executable program as soon
as its {\em Universal Search} variant has found a proof that the rewrite is {\em useful} 
according to objective (\ref{u}).
According to the Global Optimality Theorem
\cite{Schmidhuber:03gm,Schmidhuber:05icann,Schmidhuber:05gmai,Schmidhuber:05gmconscious},
such a self-rewrite is globally optimal---no local maxima!---since
the self-referential code first had to prove that it is not useful to continue the 
proof search for alternative self-rewrites.  

If there is no provably useful, globally optimal
way of rewriting $\cal S$ at all, then humans
will not find one either. But if there is one,
then $\cal S$ itself can find and exploit it.  Unlike previous {\em
non}-self-referential methods based on hardwired proof searchers \cite{Hutter:04book+},
\gmn s not only boast an optimal {\em order} of complexity but can optimally
reduce (through self-changes) any slowdowns 
hidden by the $O()$-notation, provided the utility
of such speed-ups is provable at all.

Practical implementations of the \gm do not yet exist though, and
probably require a thoughtful choice of the initial axioms
and the initial proof searcher. 
In the next sections we will deal with 
already quite practical, non-optimal and non-universal, but still
rather general searchers in program space, as opposed
to the space of reactive, feedforward input / output mappings,
which still attracts the bulk of current machine learning research.

\section{Supervised Recurrent Neural Networks}
\label{rnn}

Recurrent neural networks (RNNs) are neural networks \cite{Bishop:95}
with feedback connections that are, in principle,
as powerful as any traditional computer.
There is a very simple way to see
this~\cite{Schmidhuber:90thesis,Schmidhuber:93habil}:
a traditional microprocessor may be viewed as
a very sparsely connected RNN consisting of very
simple neurons implementing nonlinear
AND and NAND gates, etc.
Compare~\cite{siegelmann91turing}  for a more complex
argument.
Early RNNs~\cite{Hopfield:82,Almeida:87,Pineda:89}
did not exploit this potential power because
they were limited to static inputs.  However,
most interesting tasks involve processing sequences that consist of
continually varying inputs.
Examples include robot control, speech recognition,
music composition, attentive vision, and numerous
others.

The initial RNN enthusiasm of the 1980s and early 90s was fueled by the obvious
theoretical advantages of RNNs: unlike feedforward neural networks (FNNs)
\cite{Werbos:74,RumelMc:86}, Support Vector Machines (SVMs),
and related approaches~\cite{Vapnik:95}, RNNs have an internal state which is
essential for many tasks involving program learning and sequence learning.
And unlike in Hidden Markov Models (HMMs)
\cite{young2002htk}, internal states can take on
continuous values, and the influence of these states can persist for many timesteps.
This allows RNN to solve tasks that are impossible for HMMs, such as
learning the rules of certain  context-free languages
\cite{Gers:01ieeetnn}.

Practitioners of the early years,
however, have been sobered by unsuccessful attempts to apply
RNNs to important real world problems
that require sequential processing of information. The first RNNs simply
did not work very well, and their operation was poorly understood
since it is inherently more complex than that of FNNs. FNNs
fit neatly into the framework of traditional statistics and information
theory~\cite{Shannon:48}, while RNNs require additional
insights, e.g., from theoretical computer science 
and {\em algorithmic} information  
theory~\cite{LiVitanyi:97,Schmidhuber:02ijfcs}.

Fortunately, recent advances have overcome the major drawbacks of traditional RNNs. 
That's why RNNs are currently experiencing a second wave of attention.
New architectures, learning algorithms,
and a better understanding of RNN
behavior have allowed
RNNs to learn many previously unlearnable tasks. 
RNN optimists are claiming that we are at the beginning 
of a {\em ``RNNaissance''}~\cite{Schmidhuber:03rnnaissance,Schmidhuber:04learningrobots},
and that soon we will see more and more applications of the new RNNs.

Sequence-processing, supervised, gradient-based RNNs
were pioneered by Werbos \cite{Werbos:88gasmarket},
Williams \cite{Williams:89}, and
Robinson \& Fallside \cite{RobinsonFallside:87tr};
compare Pearlmutter's survey~\cite{Pearlmutter:95}.
The basic idea is:
a teacher-given training set
contains example sequences of inputs
and desired outputs or targets $d_k(t)$ for
certain neurons $y_k$ at certain times $t$.
If the $i$-th neuron $y_i$ is an input neuron,
then its real-valued activation $y_i(t)$
at any given discrete time step $t=1,2, \ldots$
is set by the environment.  Otherwise $y_i(t)$ is
a typically nonlinear function $f(\cdot)$ of
the $y_k(t-1)$ of all neurons $y_k$ connected to $y_i$,
where by default $y_i(0)=0$ for all $i$.
By choosing an appropriate $f(\cdot)$, we make the network dynamics
differentiable, e.g., $y_i(t)=arctan(\sum_k(w_{ik}y_k(t-1))$,
where $w_{ik}$ is the real-valued weight on the connection
from $y_k$ to $y_i$.
Then we use gradient descent to change the weights such
that they minimize an objective function $E$ reflecting the 
differences between actual and desired output sequences.
Popular gradient descent algorithms for computing
weight changes
$\triangle w_{ik} \sim \frac{\partial E}{\partial w_{ik}}$
include
{\em Back-Propagation Through Time} (BPTT)
\cite{Werbos:88gasmarket,Williams:89}
and {\em Real-Time Recurrent Learning} (RTRL)
\cite{RobinsonFallside:87tr,Williams:89} and
mixtures thereof
\cite{Williams:89,Schmidhuber:92ncn3}.

The nice thing about
%program search
gradient-based
RNNs is that we can {\bf differentiate our wishes with
respect to programs,} e.g.,
\cite{Schmidhuber:90sandiego,Schmidhuber:90thesis,Schmidhuber:93habil}.
The set of possible weight matrices represents
a continuous space of programs, and the objective
function $E$ represents our desire to minimize
the difference between what the network does and
what it should do. The gradient $\frac{\partial E}{\partial w}$
(were $w$ is the complete weight matrix) tells us how
to change the current program such that it will
be better at fulfilling our wish.

Typical RNNs trained by BPTT and RTRL and other previous approaches~\cite{Mozer:89focus,Mozer:92nips,Sun:93,Lin:96,Plate:93,Ring:93,Vries:91,Pearlmutter:95,Pearlmutter:89,Elman:88,Fahlman:91,Williams:89,Schmidhuber:92ncn3,Schmidhuber:92ncfastweights,WilliamsPeng:90,Watrous:92,Miller:93,Puskorius:94},
however, cannot learn to look far back into the past. Their problems were 
first rigorously analyzed in 1991 on the author's 
RNN long time lag project ~\cite{Hochreiter:91,Hochreiter:01book};
also compare Werbos' concept
of ``sticky neurons'' \cite{Werbos:92sticky}. The error signals
``flowing backwards in time'' tend to either (1) blow up or (2)
vanish: the temporal evolution of the back-propagated error
exponentially depends on the weight magnitudes.  Case (1) may lead to
oscillating weights. In case (2), learning to bridge long time lags
takes a prohibitive amount of time, or does not work at all.
So then why bother with RNNs at all? For short time lags
we could instead use short
time-windows combined with non-recurrent approaches such as
multi-layer perceptrons~\cite{Bishop:95}
or better {\em Support Vector Machines} SVMs~\cite{Vapnik:95}.

An RNN called "Long Short-Term Memory" or LSTM (figure S1)~\cite{Hochreiter:97lstm}
overcomes the fundamental problems of traditional RNNs, and efficiently
learns to solve many previously unlearnable tasks involving:
Recognition of temporally extended patterns in noisy input sequences
\cite{Hochreiter:97lstm,Gers:2000nc};
Recognition of the temporal order of widely separated
events in noisy input streams
\cite{Hochreiter:97lstm};
Extraction of information conveyed by the
temporal distance between events
\cite{Gers:2000b};
Stable generation of precisely timed rhythms,
smooth and non-smooth periodic trajectories;
Robust storage of high-precision real
numbers across extended time intervals;
Arithmetic operations on continuous input streams
\cite{Hochreiter:97lstm,Gers:2000d}.
This made possible the numerous applications
described further below.

We found~\cite{Hochreiter:97lstm,Gers:2000nc}
that LSTM clearly outperforms
previous RNNs on tasks that
require learning the rules of regular languages (RLs) describable
by deterministic finite state automata (DFA)
\cite{casey96dynamics,siegelmann93foundations,Blair+Pollack:1997nc,kalinke98computation,zeng94discrete}, both in terms of reliability and speed.
In particular, problems that are hardly ever solved by
standard RNNs, are solved by LSTM in nearly 100\% of all trials,
LSTM's superiority also carries over to tasks involving
context free languages (CFLs)
such as those discussed in the RNN literature
\cite{Sun93:abRNN,wiles95learning,steijvers96recurrent,tonkes97learning,Rodriguez:1999CS,Rodriguez+Wiles:1998:nips10}.
Their recognition requires the functional equivalent of a stack.
Some previous RNNs even failed to learn small
CFL training sets~\cite{Rodriguez+Wiles:1998:nips10}.
Those that did not
and those that even learned small CSL training sets~\cite{Rodriguez:1999CS,boden00context-free}
failed to extract the
general rules, and did not generalize
well on substantially larger test sets.
In contrast, LSTM generalizes extremely well.
It requires only the 30 shortest exemplars
($n \leq 10$) of the context sensitive language $a^nb^nc^n$ to
correctly predict the possible continuations of sequence prefixes
for $n$ up to 1000 and more.

Kalman filters can further improve LSTM's performance~\cite{Perez:02}.
A combination of LSTM and the decoupled extended Kalman filter
learned to deal correctly with values of $n$ up to 10 million and more.
That is, after training the network was able to
read sequences of 30,000,000 symbols and more,
one symbol at a time, and
finally detect the subtle differences between
{\em legal} strings such as
$a^{10,000,000}b^{10,000,000}c^{10,000,000}$
and
very similar but {\em illegal} strings such as
$a^{10,000,000}b^{9,999,999}c^{10,000,000}$. 
This illustrates that LSTM networks can work in
an extremely precise and robust fashion across
very long time lags.

Speech recognition is still dominated by 
Hidden Markov Models (HMMs), e.g., \cite{bourlard+morgan:1994}.
HMMs and other graphical models
such as Dynamic Bayesian Networks (DBN)
{\em do} have internal states that can be used to model memories of
previously seen inputs. Under certain circumstances
they allow for learning the prediction of
sequences of labels from unsegmented input streams.
For example, an unsegmented acoustic
signal can be transcribed into a sequence of words or phonemes.
HMMs are well-suited for noisy inputs and are invariant 
to non-linear temporal stretching---they do not care for the difference
between slow and fast versions of a given spoken word.
At least in theory, however, RNNs could offer the following advantages:
Due to their discrete nature, HMMs either have to discard real-valued
information about timing, rhythm, prosody, etc., or use numerous
hidden states to encode such potentially relevant information in
discretized fashion.  RNNs, however, can naturally use their
real-valued activations to encode it compactly.  Furthermore, in
order to make HMM training tractable, it is necessary to assume
that successive inputs are independent of one another. In most
interesting sequence learning tasks, such as speech recognition,
this is manifestly untrue. Because RNNs model the conditional
probabilities of class occupancy directly, and do not model the
input sequence, they do not require this assumption. RNN classifications
are in principle conditioned on the entire input sequence.  Finally,
HMMs cannot even model the rules of context-free languages, while RNNs can
\cite{Gers:2000nc,Gers:01ieeetnn,Gers:02jmlr,Schmidhuber:02nc,Perez:03}.

LSTM recurrent networks
were trained from scratch on utterances from the
TIDIGITS speech database.
It was found \cite{beringer:05icann,graves05retraining,Graves:06icml}
that LSTM obtains results comparable to HMM based systems.
A series of experiments on disjoint subsets of the database
demonstrated that previous experience greatly reduces the network's
training time, as well as increasing its accuracy.
%%ABG Since this formof adaptation would be impossible with HMMs, it was argued that LSTM is a
It was therefore argued that LSTM is a
promising tool for applications requiring either rapid cross corpus
adaptation or continually expanding datasets.
Substantial promise also lies in LSTM-HMM hybrids
that try to combine the best of both worlds, inspired by
Robinson's hybrids based on traditional RNNs~\cite{robinson:1994}.
Recently we showed~\cite{graves:05ijcnn,graves:05nn}
that LSTM learns framewise phoneme
classification much faster than previous RNNs.
Best results were obtained with a bi-directional variant of
LSTM that classifies any part of a sequence
by taking into account its entire past and future context.

Gradient-based LSTM also has been used to identify protein sequence motifs
that contribute to classification~\cite{hochreiter:snowbird}.
Protein classification is important for
extracting binding or active sites on a protein in order
to develop new drugs, and in determining 3D protein folding features
that can provide a better understanding of diseases resulting
from protein misfolding.

Sometimes gradient information is of little use due to
rough error surfaces with numerous local minima.
For such cases,
we have recently introduced a new, evolutionary/gradient-descent hybrid
method for training LSTM and other RNNs called
Evolino~\cite{Schmidhuber:05ijcai,Wierstra:05geccoevolino,Schmidhuber:06esann,Schmidhuber:06nc}.
Evolino evolves weights to the
nonlinear, hidden nodes of RNNs
while computing optimal linear mappings from hidden state to output,
using methods such as pseudo-inverse-based  linear regression
\cite{penrose:pseudo}
or support vector machine-like quadratic programming~\cite{Vapnik:95},
depending on the notion of optimality employed.
Evolino-based LSTM  can solve
tasks that Echo State networks \cite{Jaeger:04}
cannot, and achieves higher accuracy in certain continuous
function generation tasks than
gradient-based LSTM, as well as other conventional gradient descent RNNs.
However, for several problems requiring large networks with numerous
learnable weights, gradient-based LSTM was superior to
Evolino-based LSTM.

\section{Reinforcement-Learning / Evolving Recurrent Neural Networks}
\label{rlrnn}

In a certain sense,
Reinforcement Learning (RL) is more challenging than
supervised learning as above, since there is no teacher
providing desired outputs at appropriate time steps.
To solve a given problem, the learning agent itself 
must discover useful output sequences in response
to the observations.
The traditional approach to RL is best embodied by Sutton
and Barto's book \cite{Sutton:98}.  It makes strong assumptions
about the environment, such as the Markov assumption:
the current input of the agent tells it all it needs to know
about the environment. 
Then all we need to learn is some sort of reactive mapping from 
stationary inputs to outputs. This is often unrealistic.

More general approaches search a space of truly
sequence-processing programs with
temporary variables for storing previous observations.
For example, Olsson's ADATE \cite{Olsson:95} or
related approaches such as {\em Genetic Programming (GP)} 
\cite{Cramer:85,Dickmanns:87,Schmidhuber:87}
can in principle be used to evolve such programs 
by maximizing an appropriate objective or fitness function.
{\em Probabilistic Incremental Program Evolution (PIPE)}
\cite{Salustowicz:97ecj}
is a related technique for automatic program synthesis, combining
probability vector coding of program instructions
\cite{Schmidhuber:97ssa} and
Population-Based Incremental Learning
\cite{Baluja:95} and tree-coded programs.  PIPE was 
used for learning soccer team strategies
in rather realistic simulations
\cite{Salustowicz:98mlj,Wiering:99}.

A related, rather general approach for partially observable environments
directly evolves programs for recurrent neural networks (RNN) 
with internal states, by applying evolutionary
algorithms \cite{Rechenberg:71,Schwefel:74,Holland:75} to RNN weight matrices
\cite{miller:icga89,yao:review93,yamauchi94sequential,nolfi:alife4,miglino95evolving,Sims:1994:EVC,moriarty:ml96}. 
RNN can run general programs with memory / internal states (no need
for the Markovian assumption),
but for a long time it was unclear how to efficiently
evolve their weights to solve complex RL tasks.
Recent work, however, brought progress through a focus on reducing
search spaces by co-evolving the comparatively small
weight vectors of individual recurrent neurons
\cite{gomez:ab97,gomez:ijcai99,gomez:phd,Gomez:03+,Gomez:05icann,Gomez:05geccocern}.
The powerful RNN learn to use their potential to create
memories of important events, solving numerous
RL / optimization tasks unsolvable by traditional RL methods 
\cite{Gomez:03+,Gomez:05icann,Gomez:05geccocern}.
As mentioned in the previous section, even supervised
learning can greatly profit from this approach
\cite{Schmidhuber:05ijcai,Wierstra:05geccoevolino,Schmidhuber:06esann,Schmidhuber:06nc}.

\section{Is History Converging? Again?}
Many predict 
that within a few decades there will be computers
whose raw computing power will surpass the one of a human
brain by far (e.g., \cite{Moravec:99,Kurzweil:05}).
We have argued that algorithmic advances
are keeping up with the hardware development,
pointing to new-millennium theoretical insights 
on universal problem solvers that are optimal in various mathematical senses
(thus making {\em General AI} a real formal science), as well as
practical progress in program learning through
brain-inspired recurrent neural nets
(as opposed to mere pattern association through 
traditional reactive devices).

Let us put the AI-oriented developments 
\cite{Schmidhuber:06ai}
discussed above in a broader context, and
extend the analysis of past computer science 
breakthroughs in the introduction, which
predicts that computer history will converge in 
an {\em Omega point}
or historic {\em singularity} 
$\Omega$ 
around $2040$. 

Surprisingly,
even if we go back all the way to the beginnings of modern man 
over 40,000 years ago, essential historic developments (that is, 
the subjects of the major chapters in history books) match a
a binary scale marking exponentially declining temporal intervals,
each half the size of the previous one, and even measurable 
in terms of powers of 2 multiplied by a human lifetime 
\cite{Schmidhuber:06history}
(roughly 80 years---throughout recorded history many individuals
have reached this age, although the average lifetime often
was shorter, mostly  due to high children mortality).
Using the value $\Omega =2040$,
associate an error bar of not much 
more than 10 percent with each date below:
\begin{enumerate}
\item
$\Omega - 2^9$ lifetimes: 
modern humans start colonizing
the world from Africa 
\item
$\Omega - 2^8$ lifetimes: 
bow and arrow invented; hunting revolution
\item
$\Omega - 2^7$ lifetimes: 
invention of agriculture; first
permanent settlements; beginnings of civilization
\item
$\Omega - 2^6$ lifetimes: 
first high civilizations (Sumeria, Egypt),
and the most important invention of recorded history,
namely, the one that made recorded history possible: writing 
\item
$\Omega - 2^5$ lifetimes: 
the ancient Greeks invent democracy and
lay the foundations of Western science and art and
philosophy, from algorithmic procedures and formal proofs to
anatomically perfect sculptures, harmonic music, and organized sports.
Old Testament written, 
major Asian religions founded.
High civilizations in China, origin of the first 
calculation tools, and India, origin of the zero
\item
$\Omega - 2^4$ lifetimes: 
bookprint (often called the most important invention of
the past 2000 years) invented in China. 
Islamic science and culture start spreading across 
large parts of the known world (this has sometimes been 
called the most important event between Antiquity and
the age of discoveries)
\item
$\Omega - 2^3$ lifetimes: 
the largest and most dominant 
empire ever (probably including more than half of
humanity and two thirds of the world economy)
stretches across Asia from Korea all the way to Germany.
Chinese fleets and later also European vessels start
exploring the world. Gun powder and guns invented in China.
Rennaissance and Western bookprint 
(often called the most influential
invention of the past 1000 years)
and subsequent Reformation 
in Europe. Begin of the Scientific Revolution 
\item
$\Omega - 2^2$ lifetimes: 
Age of enlightenment and rational thought in Europe.
Massive progress in the sciences; first flying machines;
start of the industrial revolution based on
the first steam engines
\item
$\Omega - 2$ lifetimes: 
Second industrial revolution based on
combustion engines, cheap electricity, and modern chemistry.
Birth of modern medicine through the
germ theory of disease;
genetic and evolution theory.
European colonialism at its short-lived peak
\item
$\Omega - 1$ lifetime: 
modern post-World War II society and pop culture emerges. 
The world-wide super-exponential population explosion (mainly
due to the Haber-Bosch process \cite{Smil:99}) is at its peak.
First commercial computers and first spacecraft; 
DNA structure unveiled
\item
$\Omega - 1/2$ lifetime: 
3rd industrial revolution based on
personal computers and the World Wide Web.
A mathematical theory of universal AI emerges (see sections above) - 
will this be considered a milestone in the future?
\item
$\Omega - 1/4$ lifetime: 
This point will be reached in a few years.
See introduction
\item
...
\end{enumerate}

The following disclosure should help the
reader to take this list with a grain of salt 
though. The author, who admits being very interested in witnessing the 
Omega point, was born in 1963, and therefore perhaps should not 
expect to live long past 2040.  This may motivate him
to uncover certain historic patterns that fit his desires, while 
ignoring other patterns that do not.

Others may feel attracted by the same trap.
For example, Kurzweil \cite{Kurzweil:05} identifies
exponential speedups in sequences of historic paradigm shifts identified 
by various historians, to back up the hypothesis that ``the singularity
is near.'' His historians are all contemporary though, 
presumably being subject to a similar bias. People of past
ages might have held quite different views.  For example, possibly some 
historians of the year 1525 felt inclined to predict a convergence 
of history around 1540, deriving this date from an exponential 
speedup of recent breakthroughs such as
Western bookprint (around 1444), the re-discovery of America 
(48 years later), the Reformation (again 24 years later---see the pattern?), 
and other events they deemed important although today they are mostly forgotten.

In fact, could it be that such lists just reflect the human way of allocating
memory space to past events? Maybe there is a general rule for both
the individual memory of single humans and the collective memory
of entire societies and their history books: constant amounts of
memory space get allocated to exponentially larger, adjacent
time intervals further and further into the past. For example, events
that happened between 2 and 4 lifetimes ago get roughly as much
memory space as events in the previous interval of twice the size.
Presumably only a few ``important'' memories will survive
the necessary compression. Maybe that's why there has never been a
shortage of prophets predicting that the end is near -  the important
events according to one's own view of the past always seem to
accelerate exponentially.

A similar plausible type of memory decay allocates $O(1/n)$ 
memory units to all events older than $O(n)$ unit time intervals.  
This is reminiscent of a bias governed by a time-reversed
Speed Prior \cite{Schmidhuber:02colt} (Section \ref{unilearn}).  

\newpage

\bibliography{bib}
\bibliographystyle{plain}
\end{document}